\documentclass[letterpaper, 10 pt, conference]{ieeeconf}  

\IEEEoverridecommandlockouts                              

\usepackage{color}
\usepackage{soul}

\title{\LARGE \bf
   Binarized P-Network: Deep Reinforcement Learning of Robot Control from Raw Images on FPGA
}

\author{Yuki Kadokawa, Yoshihisa Tsurumine, and Takamitsu Matsubara
\thanks{
    This work was supported by JSPS KAKENHI Grant Number JP21H03522.
    All authors are with the Division of Information Science, Graduate School of Science and Technology, Nara Institute of Science and Technology, Japan: \{ kadokawa.yuki.kv3,tsurumine.yoshihisa, takam-m \} @is.naist.jp}
}

\usepackage{setspace}
\usepackage{comment}
\usepackage{subfigure}
\usepackage{booktabs}
\usepackage[pdftex]{graphicx} 
\usepackage{amsmath}
\usepackage{amsfonts}
\usepackage{algorithmic}
\usepackage{caption}
\usepackage{bm}
\usepackage[space, compress]{cite}

\usepackage[ruled,vlined]{algorithm2e}
\usepackage{algorithm2e,setspace}

\usepackage{textcomp}
\usepackage{mathcomp}

\usepackage{tabularx}
\usepackage{here}

\newcommand{\tabref}[1]{Table~\ref{#1}}
\newcommand{\equref}[1]{Eq.~(\ref{#1})}
\newcommand{\figref}[1]{Fig.~\ref{#1}}
\newcommand{\chapref}[1]{Section~\ref{#1}}
\newcommand{\algref}[1]{Algorithm~\ref{#1}}

\newcommand{\bs}[1]{\boldsymbol{#1}}

\begin{document}

\maketitle
\thispagestyle{empty}
\pagestyle{empty}

\begin{abstract}
    This paper explores a Deep Reinforcement Learning (DRL) approach for designing image-based control for edge robots to be implemented on Field Programmable Gate Arrays (FPGAs). 
    Although FPGAs are more power-efficient than CPUs and GPUs, a typical DRL method cannot be applied since they are composed of many Logic Blocks (LBs) for high-speed logical operations but low-speed real-number operations. To cope with this problem, we propose a novel DRL algorithm called Binarized P-Network (BPN), which learns image-input control policies using Binarized Convolutional Neural Networks (BCNNs). To alleviate the instability of reinforcement learning caused by a BCNN with low function approximation accuracy, our BPN adopts a robust value update scheme called Conservative Value Iteration, which is tolerant of function approximation errors. 
    We confirmed the BPN's effectiveness through applications to a visual tracking task in simulation and real-robot experiments with FPGA. 
\end{abstract}

\section{Introduction}
    A Field Programmable Gate Array (FPGA) is an integrated circuit that is designed to be programmable in proprietary optimizations by a customer or a designer after manufacturing. It is often described as {\it field-programmable}. By exploiting such field-programmable capability, FPGAs are often more power-efficient than CPUs and GPUs, which have a fixed number of available calculators that contain wasteful implementation. FPGAs are drawing much attention for such edge robots as flying and walking robots with limited battery capacity \cite{fpga-uav-obstacle-detection,fpga-robot-6leg,fpga-cpu-platform-for-robot,fpga-robot-mobile}. With this background, this paper focuses on designing a control for edge robots that can be implemented on FPGAs. 
    We tackle the inability to calculate an image-input controller in real-time and discuss this issue below as a problem of conventional methods.

    Deep Reinforcement Learning (DRL) is promising for automatically designing such a controller in a data-driven manner. DRLs can train a Neural Network (NN) to learn value functions for control policies that map from raw image observations to actions for task achievement. Their potential has been demonstrated in various fields, including arcade games and robot control \cite{alpha-go,google-picking}. 
    However, FPGAs' computational characteristics must be addressed to implement NNs learned from DRLs in FPGAs.
    Although Convolutional Neural Networks (CNNs) that can handle image input are typically used for learning value functions or control policies in DRL, FPGAs are mainly composed of Logic Blocks (LBs) that calculate logical operations at high-speed but real-number operations at low-speed.
    
    \begin{figure}[t]
        \vspace{1.9mm}
        \centering
        \includegraphics[width=0.88\columnwidth]{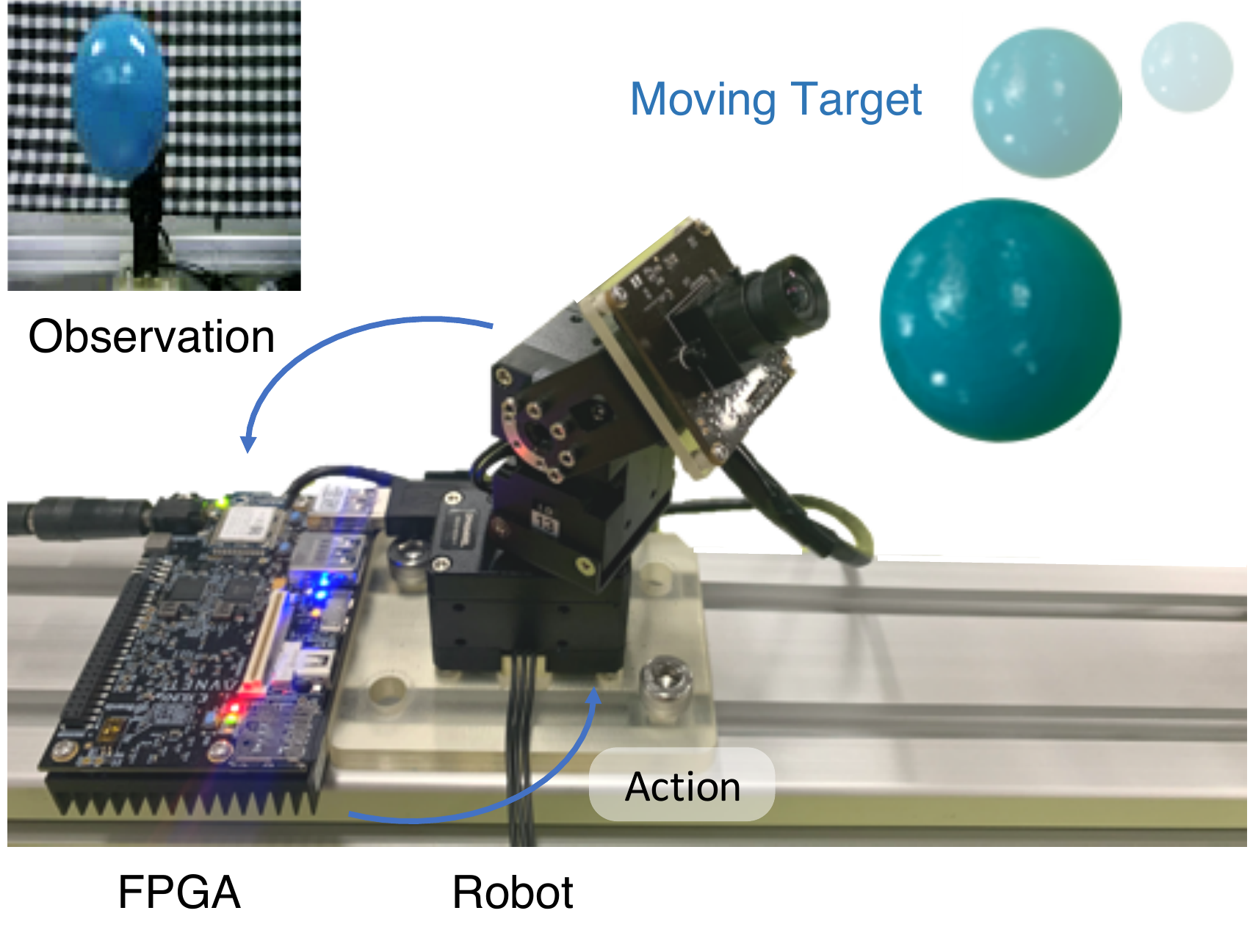}
        \caption{
        Execution of visual tracking task using proposed method: In a system consisting only of an edge FPGA and a robot, we implement a real-time control policy for image input learned by our DRL method.
        }
        \label{fig:senzai}
    \end{figure}
    
    This paper proposes Binarized P-Network (BPN) as a novel DRL algorithm that can learn image-input control policies using a Binarized Neural Network (BNN) \cite{bnn}, which is suitable for FPGA implementation. BNNs are NN models and mainly consist of logical operations. BNN can be implemented and calculated at high speed by explicitly exploiting the FPGAs' LB to calculate the network's logical operations \cite{finn,BinaryDAD-Net}. 
    However, since the approximation accuracy of the continuous function of BCNN is much lower than standard CNNs, accurately learning value functions in DRL is challenging\cite{BNN-AutoEncoder-Robot}.
    To alleviate the instability of reinforcement learning caused by using a BCNN with low function approximation accuracy, our BPN adopts a robust value update scheme, Conservative Value Iteration, which is tolerant of function approximation errors \cite{cvi}. 
    The learning procedure loops two steps: (1) The FPGA executes the policy and collects datasets. (2) The server updates the policy.
    BPN's effectiveness is validated through an application to an arm-reaching task and a visual tracking task in simulation. Moreover, we applied BPN to an object tracking task in a real-robot experiment (\figref{fig:senzai}) to learn the control policy in a real-robot environment using an FPGA. 
    
    The following are this paper's main contributions: 1) Proposed BPN, a new DRL method using a BCNN that is suitable for FPGA implementation; 2) Achieved real-time image-based robot control using BPNs.

\section{Related Works}
    Repeatedly, FPGAs are mainly composed of LBs that calculate logical operations at high-speed but real-number operations at low-speed.
    To address this problem, a naive approach uses a remote server with rich computational resources to learn and execute policies.
    However, even in a stable communication environment between the server and the edge robot, there is a considerable latency in sending and receiving sensor information and control inputs.
    This latency cannot be ignored in real-time control. 
    In addition, communication data loss may occur.
    Therefore, controlling an edge robot via a server is problematic from stability and speed of communication.
    Also, using edge-CPU may be thought helpful for executing policies fast in the edge-robot, but it is not suitable because the CNNs calculation by the edge-CPU is slow.
    Thus, previous studies have proposed the following two approaches.
    
    \textbf{Learning on FPGA \& Inference on FPGA:}
        Su et al. proposed implementing the entire flow of DRL algorithms on FPGAs, which can be applied as an approach that learns control policies through direct interaction between FPGAs and edge robots \cite{fpga-drl-all-fpga,fpga-rl-osl}. In this approach, since the robot and FPGA can communicate without the network environment, the communication delay's influence is negligible. However, FPGA cannot quickly calculate the learning algorithm and control policies. Thus, such an approach is limited to small-scale NNs and is unavailable for CNNs with image input.
        
    \textbf{Learning on Server \& Inference on FPGA:}
        Shao et al. proposed a simulation environment on a server to learn control policies to offload the learned control policies to FPGA \cite{FPGA-drl-saverLearn-FPGAOffload-trpo,FPGA-robot-DRL-AllSim-actor-critic}. Unfortunately, this proposal is again limited to small-scale NNs due to the slow calculation speed of the offloading policies in LBs on FPGAs. 
        Learning performance is also likely to be poor due to modeling errors between the simulation and real-robot environments. 
    
    Based on the above, to realize a real-time controller with image input for edge robots on FPGAs, a system's server must remotely communicate with the robot to learn control policies, as in Shao et al. Moreover, a novel framework must be considered that can more effectively use LBs so that the learned policies can be computed in real-time on FPGAs. The BPN proposed in this paper addresses this challenge.

\section{Preliminaries}
    \label{prepare}
    
    \subsection{Reinforcement Learning}
        Reinforcement learning (RL), which optimizes an agent's actions in an environmental model that follows the Markov Decision Process (MDP),
        has five components: $(\mathcal{S},\mathcal{A},\mathcal{T},r,\gamma)$.
        $\mathcal{S}$ is the set of observations that can be obtained from the environment, and $\mathcal{A}$ is the set of selectable actions.
        $\mathcal{T}^a_{s s'}$ is the probability of transitioning to observation $s' \in \mathcal{S}$ when action $a \in  \mathcal{A}$ is chosen in observation $s \in \mathcal{S}$. 
        The reward for making the transition is represented by $r^a_{s s'}$, and $\gamma \in [0,1)$ is the discount factor.
        Policy $\pi(a|s)$ is the probability of choosing action $a$ in the case of observation $s$.
        State value function $V^{\pi}$ is defined as \equref{V_function} as the evaluation criterion for policy $\pi$ at each observation $s$: 
        \begin{eqnarray}
            \label{V_function}
            \begin{aligned}
                V^{\pi}(s)={\mathbb{E}}_{\pi, T}\bigg[\sum_{\substack{t = 0}}^{\infty} \gamma^{t} r_{s_{t}} \bigg| s_{0}= s \bigg] ,
            \end{aligned}
        \end{eqnarray}
        where $r_{s_{t}}=\sum_{\substack{a \in \mathcal{A} \\ s' \in \mathcal{S}}}\pi(a|s_t)\mathcal{T}^a_{s_t s'}r^a_{s_t s'}$.
        The RL goal is to find optimal policy $\pi^{*}$ that satisfies the Bellman equation:
        \begin{eqnarray}
            \label{V_Bellman}
            \begin{aligned}
                V^{*}(s) = \displaystyle\max_{\pi} \sum_{\substack{a \in \mathcal{A} \\ s' \in \mathcal{S}}} \pi(a|s) \mathcal{T}_{ss'}^{a} \big(r_{ss'}^{a} + \gamma V^{*}(s')\big) , 
            \end{aligned}
        \end{eqnarray}
        where $V^{*}(s)$ is the optimal state value function.
        To evaluate policies based not only on observations $s$ but also actions $a$, the optimal action value function is defined:
        \begin{eqnarray}
            \label{Q_Bellman}
            \begin{aligned}
                Q^{*}(s, a) \hspace{-0.05cm}=\hspace{-0.05cm} \displaystyle\max_{\pi} \hspace{-0.05cm} \sum_{s' \in \mathcal{S}}\mathcal{T}_{ss'}^{a}\big(r_{ss'}^{a} \hspace{-0.05cm}+\hspace{-0.05cm}  \gamma \hspace{-0.1cm}  \sum_{a' \in \mathcal{A}} \hspace{-0.05cm} \pi(a'|s')Q^{*}(s', a')\big) , \hspace{-0.25cm}
            \end{aligned}
        \end{eqnarray}
        where $Q^{*}(s)$ is an optimal Q function.

    \subsection{Conservative Value Iteration}
        
        Conservative Value Iteration (CVI) is an RL method based on a value function that is robust to function approximation errors \cite{cvi}.
        CVI uses current policy $\pi$ and baseline policy $\bar{\pi}$ and adds constraint $i_{\bar{\pi}}^{\pi}$ to the learning to maintain moderate policy updates.
        CVI's goal is to find policy $\pi$ that satisfies the following modified Bellman equations:
        \begin{eqnarray}
            \label{eq:CVI-objective}
            V^{*}(s) = \displaystyle\max_{\pi} \! \sum_{\substack{a \in \mathcal{A} \\ s' \in \mathcal{S}}} \pi(a|s) \bigg[ \mathcal{T}_{ss'}^{a} \big(r_{ss'}^{a} \! + \! \gamma V^{*}(s')\big) \!+\! i_{\bar{\pi}}^{\pi}(s)\bigg] , \!
        \end{eqnarray}
        \begin{eqnarray}
            \label{eq:CVI-objective-Constraint}
            i_{\bar{\pi}}^{\pi}(s) \!=\! \sum_{a \in \mathcal{A}} \! \pi(a|s) \bigg[\! -\frac{1\!-\!\alpha}{\beta} \log \pi(a|s) \! - \frac{\alpha}{\beta} \log \frac{\pi(a|s)}{\bar{\pi}(a|s)} \bigg] , \!
        \end{eqnarray}
        where $\alpha \in [0,1]$ and $\beta \in (0,\infty)$ are hyperparameters.
        In contrast to the Q-function, the action preference function, denoted by $P$, is defined:
        \begin{equation}
            \begin{split}
                \label{eq:CVI-action-value-function} 
                P^{\pi}(s, a) =& \sum_{\substack{s' \in \mathcal{S}}} \mathcal{T}_{ss'}^{a} (r_{ss'}^{a} + \gamma \sum_{a \in \mathcal{A}} \pi(a|s) V^{\pi}(s', a')) \\
                & + \frac{\alpha}{\beta}\log \pi(a|s). 
            \end{split}
        \end{equation}
        To find optimal policy $\pi^*$ that maximizes \equref{eq:CVI-action-value-function}, the update rule of action preference $P$ is defined:
        \begin{equation}
            \begin{split}
                \label{eq:CVI-update}
                P_{k+1}(s, a) \gets & r_{ss'}^{a}+ \gamma (m_{\beta} P_{k})(s') + \mathcal{G}(s,a) , \\
                \mathcal{G}(s,a) & = \alpha \bigg(P_{k}(s, a) - (m_{\beta} P_{k})(s)\bigg) , 
            \end{split}
        \end{equation}
        \vspace{-1mm}
        \begin{equation}
            \label{eq:mellowmax}
             \left( m_{\beta} P \right) (s) = \frac{1}{\beta} \log \left( \frac{1}{|\mathcal{A}|} \sum_{a \in A}\exp \left(\beta P(s,a)\right) \right) , 
        \end{equation}
        where $|\mathcal{A}|$ is the number of selectable actions.
        The policy is given as follows: 
        \begin{equation}
            \label{eq:policy}
            \pi_{k}(s,a) = \frac{\exp \left(\beta P_{k}(s,a)\right)}{\sum_{b\in A} \exp \left(\beta P_{k}(s,b)\right)}. 
        \end{equation}
        $\mathcal{G}(s,a)$ in \equref{eq:CVI-update} is the Gap Increasing Operator (GIO) \cite{cvi} that amplifies the differences between the maximum value and others. Therefore, it makes the resulting policy for choosing optimal action robust against function approximation errors \cite{cvi}. 
        We refer to $\alpha$ as the GIO coefficient.
        When the $\alpha$ is higher, the robustness to the function approximation errors is higher.
        Also, the $\beta$ controls learning convergence.
        When the $\beta$ is higher, the learning convergence is faster.
        
        When the GIO coefficient is $\alpha=1$, it is theoretically equivalent to Dynamix Policy Programming (DPP) \cite{DPP-origin}, which has been used in previous studies to learn robot control policies and improved sample efficiency \cite{dpn}. 
        Moreover, CVI is nearly equivalent to Q-learning when the parameters are set as $\alpha=0$ and $\beta=\infty$ \cite{cvi}.
        The parameters mean that DQN updates the value function in greedy.
        Thus, the learning performance of DQN becomes degraded when the function approximation accuracy is low since DQN is sensitive to the function approximation errors. 
        It means that CVI with high $\alpha$ and certain $\beta$ is suitable for learning a policy calculated in low accuracy of function approximation. 
        
    \subsection{Binarized Neural Network}
        \label{bnn}
        
        This section briefly summarizes Binarized Neural Networks (BNNs), which are neural networks with binary weights and run-time activations. 
        Assuming that the dimensions of the input and output vectors in each layer of the BNN are $N$ and $M$, hierarchical functions output $\bs{y}\in\mathbb{R}^{M}$ from input $\bs{x}\in\mathbb{R}^{N}$.
        Each layer consists of a Fully-Connected Layer (FCL) and an activation function.
        Assuming that the number of BNN layers is $L$, the FCL output of the $l$th layer is $\bs{o}_l=[o_{l,1},o_{l,2},\dots,o_{l,M}]^{\rm{T}}\in\mathbb{R}^M$, the output of the activation function of the $l$th layer is $\bs{x}_l=[x_{l,1},x_{l,2},\dots,x_{l,M}]^{\rm{T}} \in \mathbb{R}^M$, and the BNN's network parameters of the $l$th layer are $\bs{W}_l =[\bs{W}_{l,1}\bs{W}_{l,2}\dots \bs{W}_{l,M}] \in  {R}^{N \times M}$, $\bs{W}_{l,m} =[w_{l,m,1} \ w_{l,m,2} \dots w_{l,m,N}] \in \mathbb{R}^{N \times M}$ set to $\bs{\theta}=\{\bs{W}_1, \bs{W}_2, \dots, \bs{W}_L \}$.
        
        As a key feature of BNN, binarized function $\text{Sign}:\mathbb{R}\rightarrow \{ -1,1 \}$ in \equref{eq:binarize} is included in FCL and in the activation functions of each layer: 
        \begin{eqnarray}
            \label{eq:binarize}
            z^b &=& {\text{Sign}}(z) = \left \{
                    \begin{array}{l}
                        +1, \;\;\;\;\;\; \text{if} \ z\geq 0 , \\
                        -1, \;\;\;\;\;\; {\text{otherwise}} , 
                    \end{array}
                \right.
        \end{eqnarray}
        where $z$ is an arbitrary real number and the value after binarization is denoted as $z^b$.
        Assuming that BNN input $\bs{x}$ and output $\bs{y}$ are $\bs{x}_0$ and $\bs{o}_L$, the operation of each layer is given below:
        \begin{subequations}
            \label{eq:bnn}
            \begin{eqnarray}
                \label{eq:bnn-fc}\hspace{-1.5em} {o}_{l,m} \hspace{0em} &=& \sum_{n=0}^{N} {\text{Sign}(w_{l,m,n}) x_{l-1,n}} \hspace{0.2em}, \\
                \label{eq:bnn-activation}\hspace{-1.5em}{x}_{l,m} \hspace{0em} &=& \text{Sign}(o_{l,m}) \hspace{0.2em}.
            \end{eqnarray}
        \end{subequations}
        In \equref{eq:bnn-fc}, since each value of parameter $\bs{W}_{l,m,n}$ can be converted to $1$ and $0$,  each weight can represented by a single bit. Therefore, by storing each element of parameter $\bs{W}_l$ in 1 bit, the model size can be compressed to 1/32 compared to single-precision, floating-point numbers.
        
        Moreover, by binarizing input $\bs{x}_{l-1}$ to each layer using the activation function in \equref{eq:bnn-activation}, the multiplication-and-accumulation (MAC) operations in \equref{eq:bnn-fc} can be replaced with the {\it XNOR} and {\it popcount} operations:
        \begin{eqnarray}
            \label{eq:bit_count} o_{l,m} = \text{Popcount}(\text{XNOR}(w_{l,m,n}^{b}, x_{l-1,n}^{b})) , 
        \end{eqnarray}
        where weights $w_{l,m,n}^{b} \in \{-1,1\}$ and input $x_{l-1,n}^{b} \in \{-1,1\}$ are converted to $w_{l,m,n}^{b} \in \{0,1\}$ or $x_{l-1,n}^{b} \in \{0,1\}$.
        The XNOR operation corresponds to the product of $w_{l,m,n}^{b}$ and $x_{l-1,n}^{b}$, and the popcount operation corresponds to counting the number of output bits from the XNOR operation \cite{finn}. 
        Therefore, FPGAs can calculate BNN at high speed since they can calculate such logical operations in LBs. 
        Binarized Convolutional Neural Networks (BCNNs) \cite{bnn} can also be constructed by simply including convolutional layers in the same way. 
        
        BNN updates parameter $\bs{\theta}$ by gradient descent, but the learning method is different from a standard NN. 
        Updating NN parameters consists of forward- and back-propagation.
        In forward-propagation, BNN-output $\bs{y}$ is obtained using binarized weights $\bs{\theta}^{b}$. On the other hand, in back-propagation, $\bs{\theta}$ is updated by back-propagation using BNN-output $\bs{y}$ with non-binarized network parameter $\bs{\theta}$ to avoid unstable learning due to discontinuity in $\bs{\theta}^b$ \cite{bnn}.

\section{Binarized P-Network}
    \label{proposed_method}
    
    \subsection{Network Architecture}
    
        \begin{figure}[t]
            \vspace{2mm}
            \centering
                \includegraphics[width=0.9\linewidth]{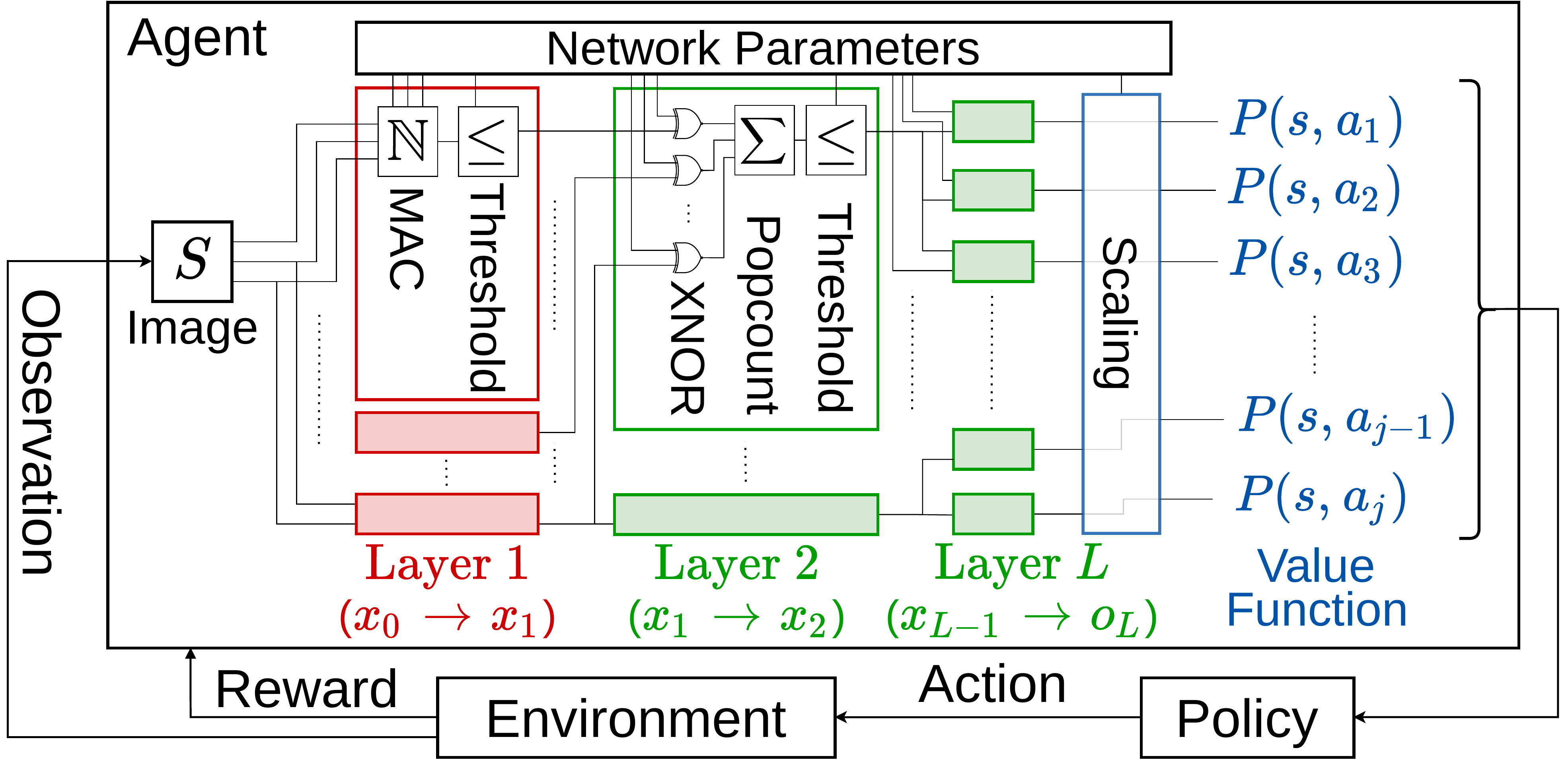}
                \caption{
                    BPN's network architecture: 
                    BPN's network calculates action preference $P(s,a)$ from an image as observation $s$ in three steps:
                    (\textbf{1}) First convolution layer extracts features from observation $s$ in MAC operations and threshold activation function.
                    (\textbf{2}) Second and subsequent layers  calculate the XNOR operations, the popcount operations, and the threshold activation functions.
                    (\textbf{3}) Last layer's outputs are scaled by $\lambda$ and used as action preference $P(s,a)$.
                }
                \label{fig:BPN_overView_en}
        \end{figure}
        
       BPN's network architecture is shown in \figref{fig:BPN_overView_en}. To alleviate the instability of reinforcement learning caused by a BCNN with low function approximation accuracy, our BPN adopts a robust value update scheme: Conservative Value Iteration. 
        Thus, based on $P(s,a)$ represented by BCNN, action $a$ is executed according to policy $\pi$ in \equref{eq:policy}.
        
        The network calculation to obtain $P(s,a)$ has three steps.  (1) The first layer, which extracts features from an image to output $\bs{x}_1$, consists of MAC operators and a threshold activation function (\equref{eq:activation-threshold}).
        MAC operators output $\bs{o}_1$, and the threshold activation function outputs $\bs{x}_1$.
        (2) In the second and subsequent layers, output $\bs{x}_2,\bs{x}_3,\dots,,\bs{x}_{L-1},\bs{o}_L$ is obtained by XNOR and popcount operations which output $\bs{o}_l$ and the threshold activation function which outputs $\bs{x}_l$.
        First half layers $\bs{o}_1,\bs{o}_2,\dots,\bs{o}_c$ are obtained by the convolutional layers \cite{finn}, and the second half layers $\bs{o}_{c+1},\bs{o}_{c+1},\dots,\bs{o}_L$ are obtained by the FCLs (\equref{eq:bit_count}).
        (3) In the last layer, FCL's output $\bs{o}_L$ is scaled by $\lambda$ and output as action preference $P(s,a)$.

    \subsection{Network Details}
        \label{chap:BPN-Network-Structure}
        
        To approximate action preference $P(s,a)$ with BCNN, which has low function approximation accuracy, we added the following three features to the network.

        \begin{algorithm}[t]
            \SetKwData{Left}{left}\SetKwData{This}{this}\SetKwData{Up}{up}
            \SetKwFunction{Union}{Union}\SetKwFunction{FindCompress}{FindCompress}
            \SetKwInOut{Input}{input}\SetKwInOut{Output}{output}
            
            \caption{Binarized P-Network}         
            \label{alg:BPN}
            
            \# Set parameters described in \tabref{table:bpn_setting} \\
            \# Initialize network weights ${\bs{\theta}}$, $\bs{\theta}^{-}$, replay memory ${\mathcal{D}}$ \\
            
            \SetKwFunction{UpPN}{PolicyUpdate}
            \SetKwFunction{DC}{DataCollect}
            \SetKwProg{Fn}{Function}{:}{\KwRet}
            
            \Fn{\DC {${\bs{\theta}}$, ${\mathcal{D}}$}}
            {
                \For{$ e = 1, 2, ..., E$}
                {
                    \For{$ t = 1, 2, ..., T$}
                    {
                        \# Take action $a_t$ with softmax policy \equref{eq:policy} based on ${P}(s_t, \mathcal{A};{\bs{\theta}}^{b})$ \\
                        \# Receive observation $s_{t\!+\!1}$, reward $r_{s_t s_{t\!+\!1}}^{a_t}$ \\
                        \# Push $\{(s_t, a_t,r_{s_t s_{t+1}}^{a_t},s_{t+1})\}$ to $\mathcal{D}$ \\
                    }
                }
                \# \Return ${\mathcal{D}}$ \\
            }
            
            \Fn{\UpPN{${\bs{\theta}}$, ${\mathcal{D}}$}}
            {
                \# Set target network $\bf{\bs{\theta}}^- = \bf{\bs{\theta}}$ \\
                
                  \For{$ c = 1, 2, ..., C$}
                  {
                    \# Set ${\mathcal{D}'}$ is index-shuffle local memory ${\mathcal{D}}$ \\
                    \For{$ k = 1, 2, ..., \rm{round}(\ |\mathcal{D}| / B\ )$}
                    {
                      \# Sample the minibatch of transition ${\mathcal{D}'}[B \! \times \! (k\!-\!1) : B \! \times \! k]$ \\
                      \# Calculate loss on \equref{eq:bpn-error} and update $\bf{\bs{\theta}}$ \\
                      }
                  }
                  \# \Return $\bf{\bs{\theta}}$ \\
            }
            
            \For{$ i = 1, 2, ..., I$}
            {
                \# ${\mathcal{D}} = \text{DataCollect}(\bf{\bs{\theta}}, {\mathcal{D}})$ \\
                \# $\bf{\bs{\theta}} = \text{PolicyUpdate}(\bf{\bs{\theta}}, {\mathcal{D}})$ \\
            }
        \end{algorithm}
        
        \begin{figure}[t]
            \vspace{1.7mm}
            \centering
                \includegraphics[width=0.92\linewidth]{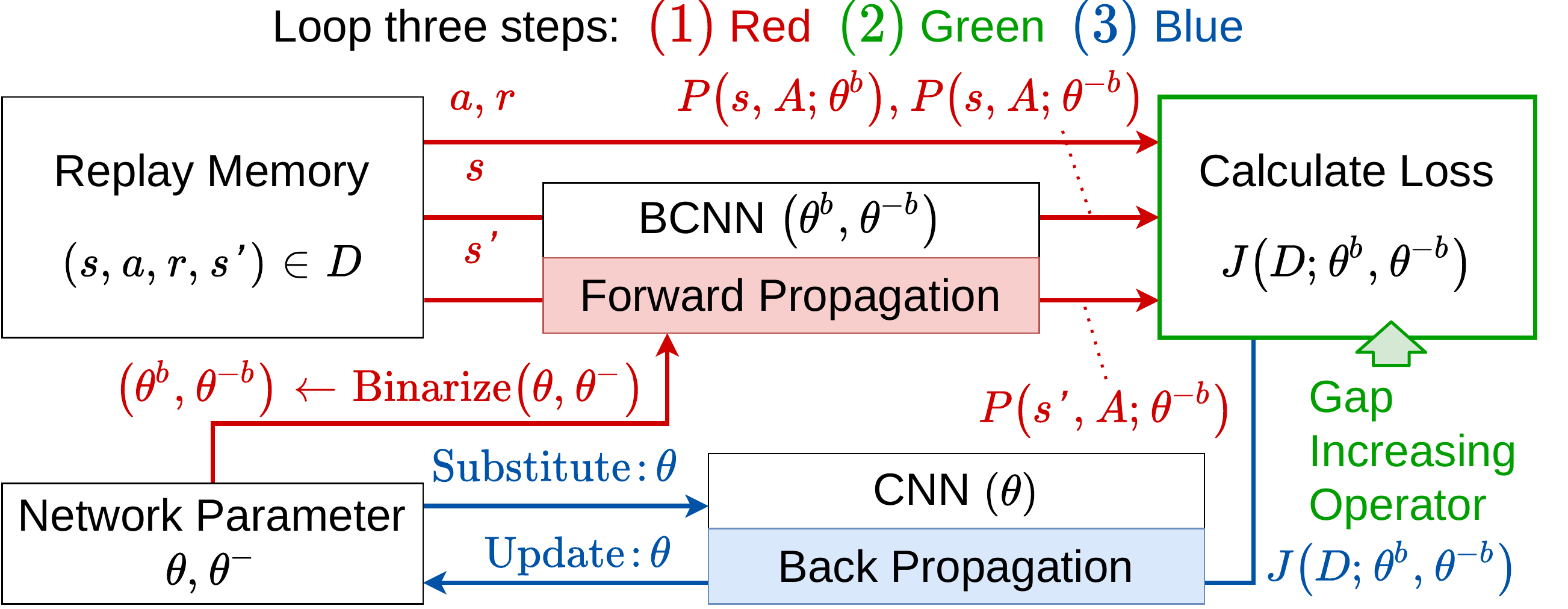}
                \caption{
                    Policy updating process of BPN: (1) Calculate action preferences, (2) Calculate loss function, (3) update network parameters. 
                }
                \label{fig:BPN-learning}
        \end{figure}
        
        \begin{table}[t]
            \vspace{1mm}
            \caption{
                    Learning parameters of BPN in two DOF manipulator reaching tasks.
                    \label{table:bpn_setting}
            }
            \vspace{-2mm}
            \begin{center}
                \begin{tabular}{@{}lp{5.5cm}llll@{}}
                    \toprule
                    \textbf{Para.} & \textbf{Meaning} & \textbf{Value}  \\ 
                    \midrule
                    $\alpha$ & GIO coefficient of CVI & 0.95 \\
                    $\beta$ & Learning speed coefficient of CVI & 1 \\
                    $\gamma$ & Discount factor of RL & 0.99 \\ 
                    $C$ & Number of epochs & 50 \\
                    $B$ & Minibatch size & 32 \\
                    $I$ & Number of iterations & 50 \\ 
                    $E$ & Number of episodes per iteration & 10 \\
                    $T$ & Number of steps per episode & 20 \\
                    $U$ & Number of iteration datasets in $\mathcal{D}$ & 3 \\ 
                    \bottomrule
                \end{tabular}
            \end{center}
        \end{table}

        \paragraph{Unbinarized observation}
            To avoid reducing the features from observation $s$, BPN doesn't binarize the first layer's input $s$. Thus, only the weights are binarized in the input convolutional layer, and the input is kept as pixel values.

        \paragraph{Batch normalization}
            BCNNs are prone to learning instability due to the binarization of weights and outputs in each layer. Thus, we added batch normalization to  every BPN layer to stabilize the learning against dynamic changes in the target value. 
            This dynamic changes of target values is unique to RL.
            Note that the calculation of batch normalization is slow on FPGAs because it has many floating-point computations.
            To speed up the calculation, the batch normalization and the activation function of \equref{eq:bnn-activation} are combined and converted into a threshold activation function:
            \begin{eqnarray}
                \label{eq:activation-threshold}
                {x}_{l,m} &=&\left \{
                        \begin{array}{l}
                            +1, \;\;\;\;\;\; \text{if} \ o_{l,m} \geq \tau_{l,m} , \\
                            -1, \;\;\;\;\;\; {\text{otherwise}} , 
                        \end{array}
                    \right.
            \end{eqnarray}
            where $\bs{\tau}_l \! = \! [\tau_{l,1},\dots,\tau_{l,M}]^{\rm{T}} \! \in \! \mathbb{N}^M$ is the $l$th layer threshold \cite{BNN_BathNormarization_theory}.

        \paragraph{Scaling network output}
            \equref{eq:bnn-fc} shows that the problem of using BCNN as a function approximator is that FCL's outputs are limited to range $[-N,N]$. 
            BPN resolves this limitation by introducing scaling parameter $\lambda$ to the last layer FCL's output $o_{L,m}$:
            \begin{eqnarray}
                \label{eq:bpn_range}
                P(s,a_m) &=& \lambda \sum_{n=0}^{N} {\text{Sign}(w_{L,m,n}) x_{L-1,n}}. 
            \end{eqnarray}
            Since it is difficult to set $\lambda$ by hand, it is learned from data.

    \subsection{Learning Process}
        \label{chap:BPN-learning}
        The learning process consists of data collection and policy update steps.
        BPN uses the target network and the replay memory, as in the DQN method \cite{dqn}.
        A target network technique uses two network parameters: P-network parameters $\bs{\theta}$ and target network parameters $\bs{\theta}^-$. 
        $\bs{\theta}^-$ decides the actions during the data collection step.
        $\bs{\theta}$ is updated in the policy update step.
        $\bs{\theta}^-$ is updated to $\bs{\theta}$ at regular intervals to stabilize the learning and moderating the frequency of the network parameter updates.
        The details of the BPN learning process are shown below and summarized in \algref{alg:BPN}.
        
        \paragraph{Data collection}
            First, target network parameters $\bs{\theta}^{-}$ are copied from P-network parameter $\bs{\theta}$.
            Then to calculate the action preferences, all the parameters in $\bs{\theta}^{-}$ are binarized to $\bs{\theta}^{-b}$ based on \equref{eq:binarize}.
            In this paper, binarized network parameters $\bs{\theta}$ are denoted as $\bs{\theta}^{b}$.
            The training datasets are then sampled based on the current policy with $\bs{\theta}^{-b}$.
            
            In the data collection step, the target network first takes observation $s$ as input and outputs action preference $P(s,\mathcal{A};\bs{\theta}^{-b})$.
            Then, based on $P(s,\mathcal{A};\bs{\theta}^{-b})$, the agent executes action $a$ based on the softmax function in \equref{eq:policy}.
            The environment transitions and outputs next observation $s'$ and reward $r$.
           $(s,a,r,s')$ pairs are added to replay memory $\mathcal{D}$ as a training dataset.

        \paragraph{Policy update}
            In the policy update step, the loss function is calculated based on dataset $\mathcal{D}$ and accumulated in the data collection step.
            \figref{fig:BPN-learning} shows how to update P-network parameters $\bs{\theta}$ in three steps.
            (1) Sets of minibatches $(s,a,r,s')$ are created from dataset $\mathcal{D}$.
            Action preferences $P(s,\mathcal{A};\bs{\theta}^b)$, $P(s',\mathcal{A};\bs{\theta}^b)$, $P(s,\mathcal{A};\bs{\theta}^{-b})$ are calculated from $s$, $s'$.
            (2) Loss function $J(\mathcal{D};\bs{\theta}^b,\bs{\theta}^{b-})$ derived from \equref{eq:CVI-update} is calculated as follow:
            \begin{equation}
                \begin{split}
                    \label{eq:bpn-error} 
                    & \hspace{-0.7em} J(\mathcal{D};\bs{\theta}^b,\bs{\theta}^{-b}) = 
                    \frac{1}{2} \left[ r^{a}_{s,s'} + \gamma (m_{\beta}P)(s';\bs{\theta}^{-b}) \right . \\
                    & \hspace{-0.7em} \left . +\alpha\left(P(s,a;\bs{\theta}^{-b}) \!-\! (m_{\beta} P)(s;\bs{\theta}^{-b})\right) - P(s,a;\bs{\theta}^b)) \right]^2. 
                \end{split}
            \end{equation}
            (3) Network parameters $\bs{\theta}$ are updated by back-propagation using a CNN composed of unbinarized network parameters $\bs{\theta}$, as described in \chapref{bnn}.

\section{Simulation Experiment}
    \label{ex:sim}
    In this section, we evaluated BPN's learning performance in a simulation study conducted with a Geforce RTX2080Ti GPU. As a comparison, we also evaluated the Binarized Q-Network (BQN) performance, which is a modified DQN with binarization for both the weights and outputs of every layer. Note that BQN is different from Binary Q-Network \cite{binary-Q-network}, which binarizes only the weights and cannot be implemented in FPGA. 
    As shown in \equref{eq:bpn_range}, the accuracy of the BPN output depends on the number of nodes $N$ in the output layer. Thus, we verify that BPN can learn a policy robustly against a variation of function approximation accuracy due to the change in the number of nodes $N$.
    
    \begin{figure}[]
        \vspace{2mm}
        \centering
        \begin{tabular}{c}
            \hspace{3mm}
            \begin{minipage}{0.5\linewidth}
                \subfigure[Reaching]{
                \label{fig:sim_all-reaching}
                \includegraphics[keepaspectratio, height=2cm, angle=0]
                            {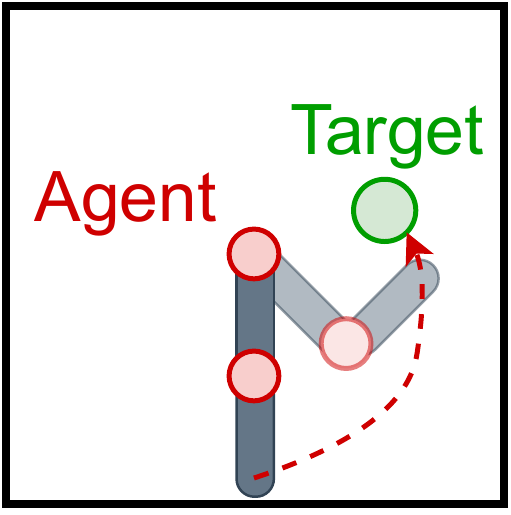}}
            \end{minipage} 
            \begin{minipage}{0.5\linewidth\hspace{-0.5cm}}
                \subfigure[Tracking]{
                \label{fig:sim_all-trucking}
                \includegraphics[keepaspectratio, height=2cm, angle=0]
                            {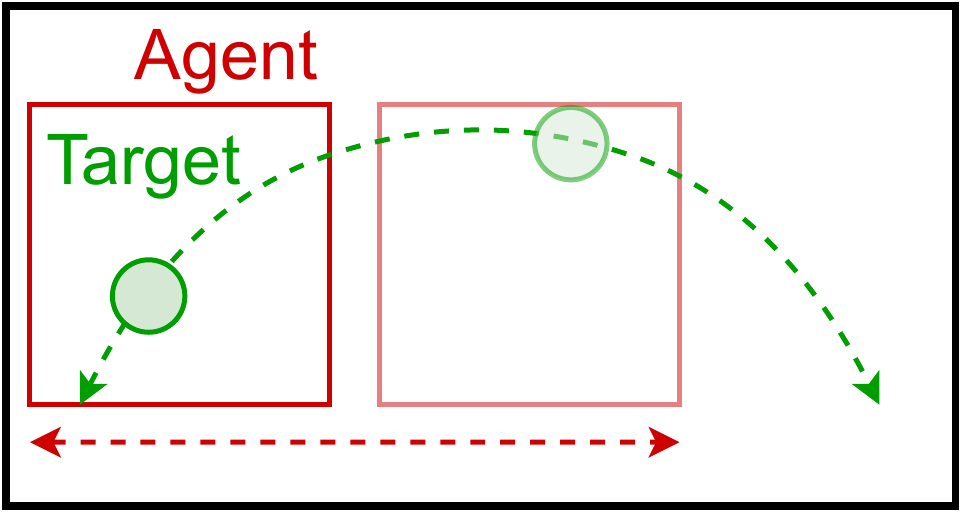}}
            \end{minipage}
        \end{tabular}
        \caption{
            Simulation tasks: (a) reaching and (b) tracking. 
        }
    \end{figure}

    \subsection{Settings}
    
        \subsubsection{Reaching Task}
            \label{sim:reaching}
            The target task is the 2DOF reaching task in \figref{fig:sim_all-reaching}.
            The agent rotates one joint at each step by a fixed angle.
            The target marker is fixed the entire time.
            The agent's learning goal is to match the hand coordinates with the target marker.
            The initial positions of the agent and the target are fixed.
            Let observation $s$ be a gray-scale image of $84\times84$ pixels obtained from the entire simulation environment, such as \figref{fig:sim_all-reaching}.
            Agent's action $a$ is selected from seven levels of target rotation angles: $[-90, -45, -30, 0, 30, 45, 90] (\text {degree})$.
            The number of selectable actions is $|\mathcal{A}|=2\times7=14$.
            The number of pixels in the horizontal and vertical directions of the image obtained from observation $s$ is defined as the XY coordinates.
            The robot's coordinates are $(x_{\text{agent}},y_{\text{agent}})$, and the target's coordinates are $(x_{\text{target}},y_{\text{target }})$.
            The reward is defined as $r = - \sqrt{(x_{\text{agent}}-x_{\text{target}})^2 + (y_{\text{agent}}-y_{\text{target}})^2}$.
            The network structure is consist of five layers, which are Conv(8,4,8), Conv(4,2,16), Conv(3,1,16), FC($N$), FC($|\mathcal{A}|$).
            Conv() means convolutional layer, which parameters are kernels, strides, and channels, respectively.
            FC() means full-connected layer, which parameter is nodes.
            The training parameters are described in \tabref{table:bpn_setting}.
        
        \subsubsection{Tracking Task}
            \label{sim:tracking}
            \figref{fig:sim_all-trucking} shows the experimental environment.
            The agent manipulates the red frame and learns that making the target always appears within it.
            The initial positions of the agent and the target are randomly assigned.
            The environment is represented by $120\times180$ pixels in height and width.
            The agent frame size is $84\times84$ pixels.
            To estimate the target's velocity, two consecutive frames are combined and used as observation $s \in \mathbb{R}^{6\times84\times84}$.
            The target is a circle with a 12-pixel radius and moves in an arc of a 60-pixel radius.
            Agent's action $a$ moves the frame horizontally by the specified number of pixels in one step.
            Action $a$ is selected from  $[-8,-4,2,0,2,4,8](\text{pixel})$.
            Reward $r$ is the distance between the center coordinates of the frame and the target.
            Reward calculation is identical as \chapref{sim:reaching}.
            However, if the target moves out of the frame, we treat it as a tracking failure and the end of the episode.
            The network structure is identical as \chapref{sim:reaching}.
            The difference between the training parameters and \tabref{table:bpn_setting} is $T=40$.

            \begin{figure}[t]
                \vspace{1mm}
                \centering
                \begin{tabular}{c}
                    \hspace{-0.6cm}
                    \begin{minipage}{0.5\linewidth}
                        \centering
                        \subfigure[Reaching]{
                        \label{fig:sim_result-reaching-all}
                        \includegraphics[keepaspectratio, width=0.95\linewidth, angle=0]
                                    {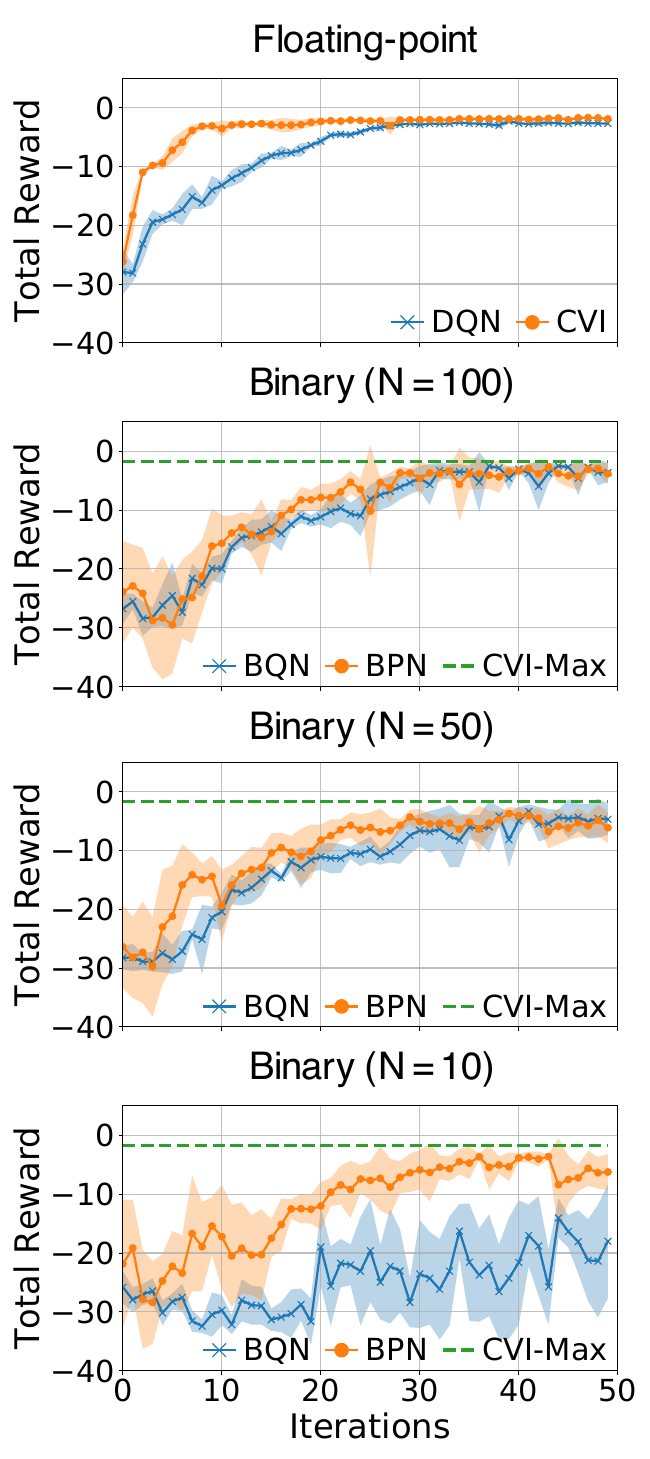}}
                    \end{minipage} 
                    \begin{minipage}{0.5\linewidth}
                        \centering
                        \subfigure[Tracking]{
                        \label{fig:sim_result-tracking-all}
                        \includegraphics[keepaspectratio, width=0.95\linewidth, angle=0]
                                    {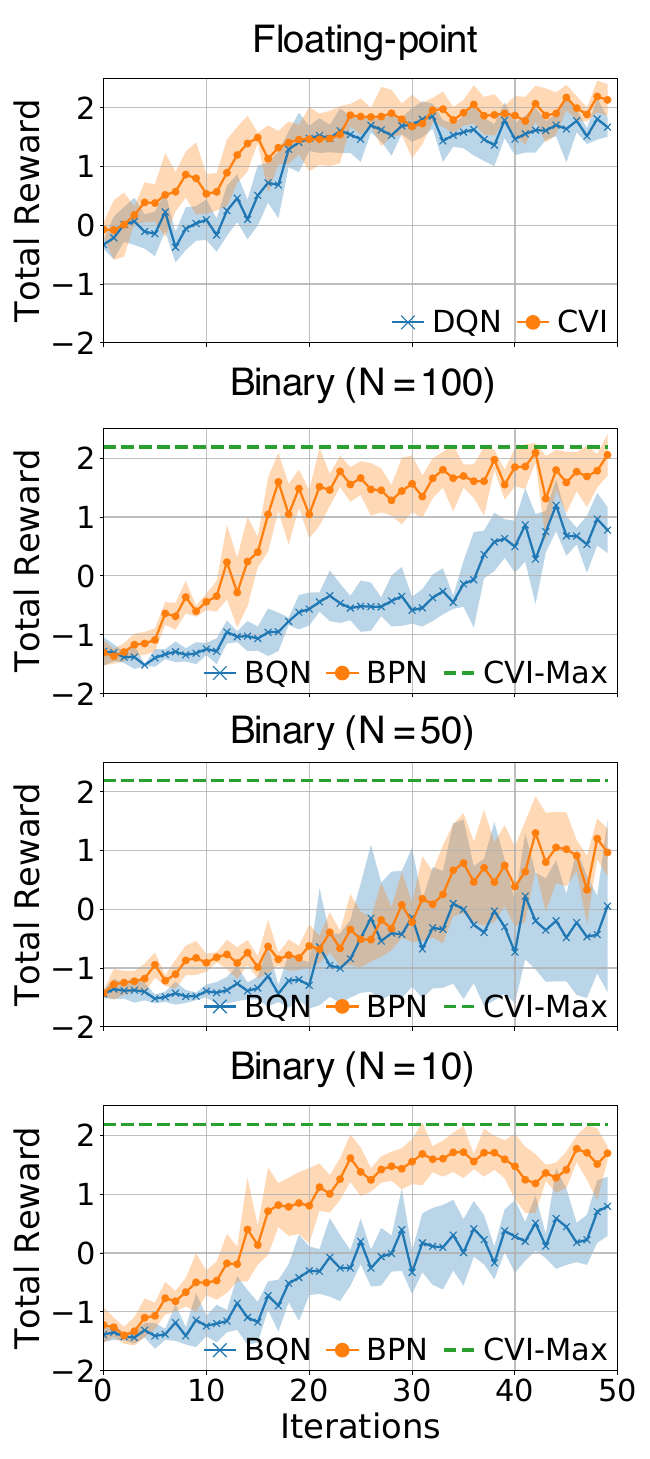}}
                    \end{minipage} 
                \end{tabular}
                \caption{
                    Learning curves of (a) reaching and (b) tracking tasks. 
                    CVI-Max indicates the maximum training reward using CVI with the floating-point NNs.
                    Each curve plots mean and variance of total reward per iteration $I$ over five experiments.
                }
                \label{fig:sim_result-all}
            \end{figure}

            \begin{figure}[t]
                \begin{center}
                    \includegraphics[width=0.90\linewidth]{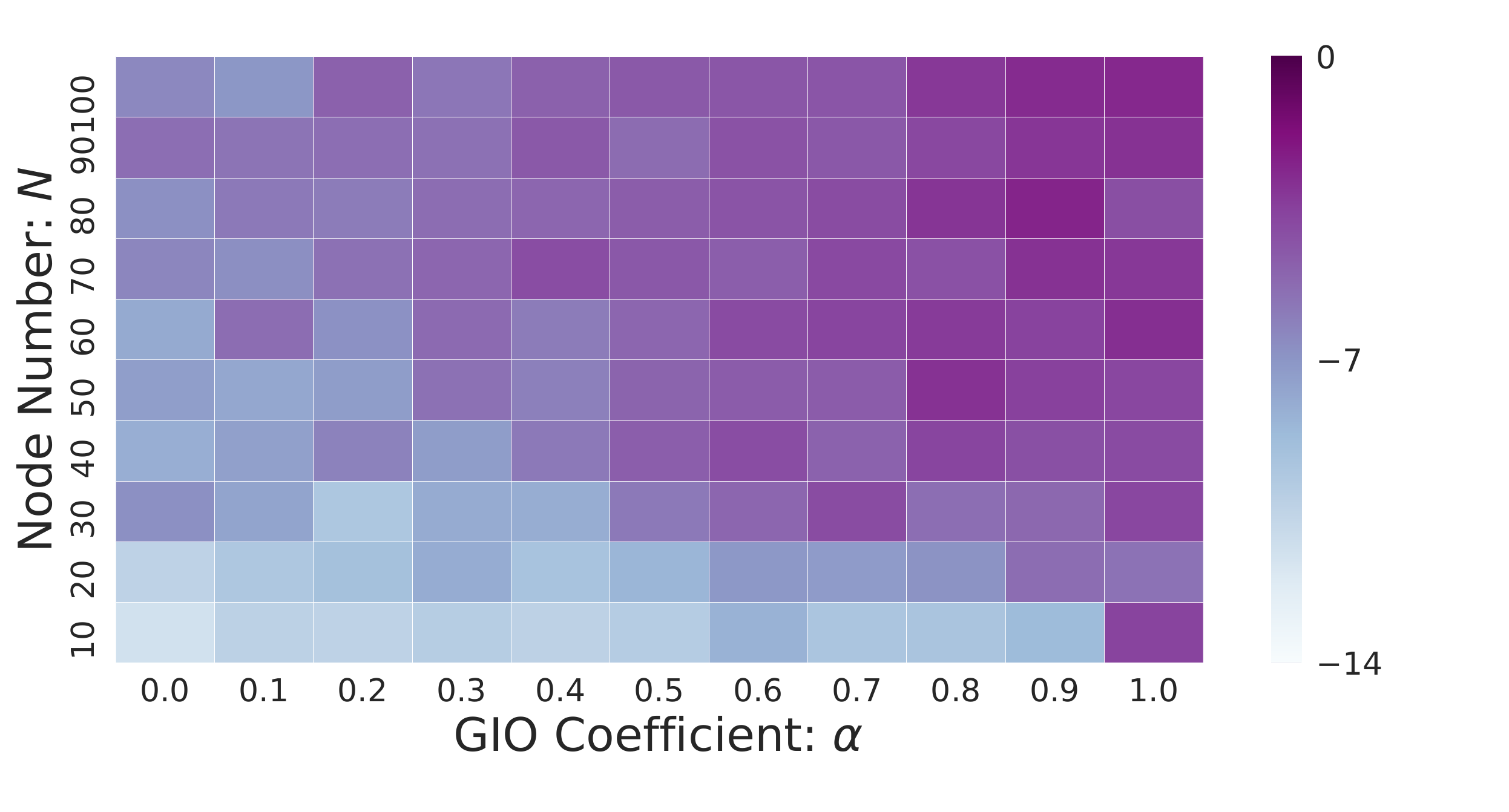}
                    \vspace{-2mm}
                    \caption{
                        Relationship between number of nodes $N$ in output layer of a function approximation network and GIO coefficients $\alpha$.
                        Each value in heat map is maximum value in total reward's learning curve averaged over five experiments.
                    }
                    \label{fig:heatmap}
                \end{center}
            \end{figure}

    \subsection{Results}
        The learning results of reaching task and tracking task are shown in \figref{fig:sim_result-reaching-all}, \figref{fig:sim_result-tracking-all}, respectively. 
        In all simulation tasks, learning performance, such as, training stability, sample efficiency, maximum total reward, is decreased when the function approximation accuracy is reduced.
        Compared to BQN, the proposed method, BPN, mitigates the decrease of performance. 
        Compared to the reaching task, BPN's learning performance in the tracking task remains high, although BQN's performance suffers. These results seem reasonable since the tracking task is more difficult than the reaching task because the initial positions of the agent and the target are randomly assigned.
        \figref{fig:heatmap} shows the relationship between node number $N$ and GIO coefficient $\alpha$.
        We confirmed that the higher GIO coefficient $\alpha$ is, the more robustly the BPN can learn against a decrease in node number $N$.
        This result is consistent with the property of the GIO operator, where the higher the GIO coefficient $\alpha$ is, the more robust it is to function approximation errors.

\section{Real-Robot Experiment}
    \label{ex:real}
    This section shows the structure of a DRL system using FPGA and robots to learn control policies.
    Using the DRL system, we apply BPN and BQN to a real-robot object tracking task and verify the learning performance.
    We also analyzed the calculation speed in the FPGA implementation to confirm that BPN is suitable for real-time control.

    \subsection{Learning System for DRL with FPGAs}
        \label{chap:real_ex_sistem}
        The policy updates of BPN were conducted on the GPU server since the BPs of NNs in policy updates require many floating-point operations, and FPGAs do not have enough LBs to calculate them.
        Hence, the GPU server calculates the policy updates; the FPGA calculates only the policy executions.
        
\begin{figure}[t]
            \vspace{1.7mm}
            \begin{center}
                \includegraphics[width=0.8\linewidth]{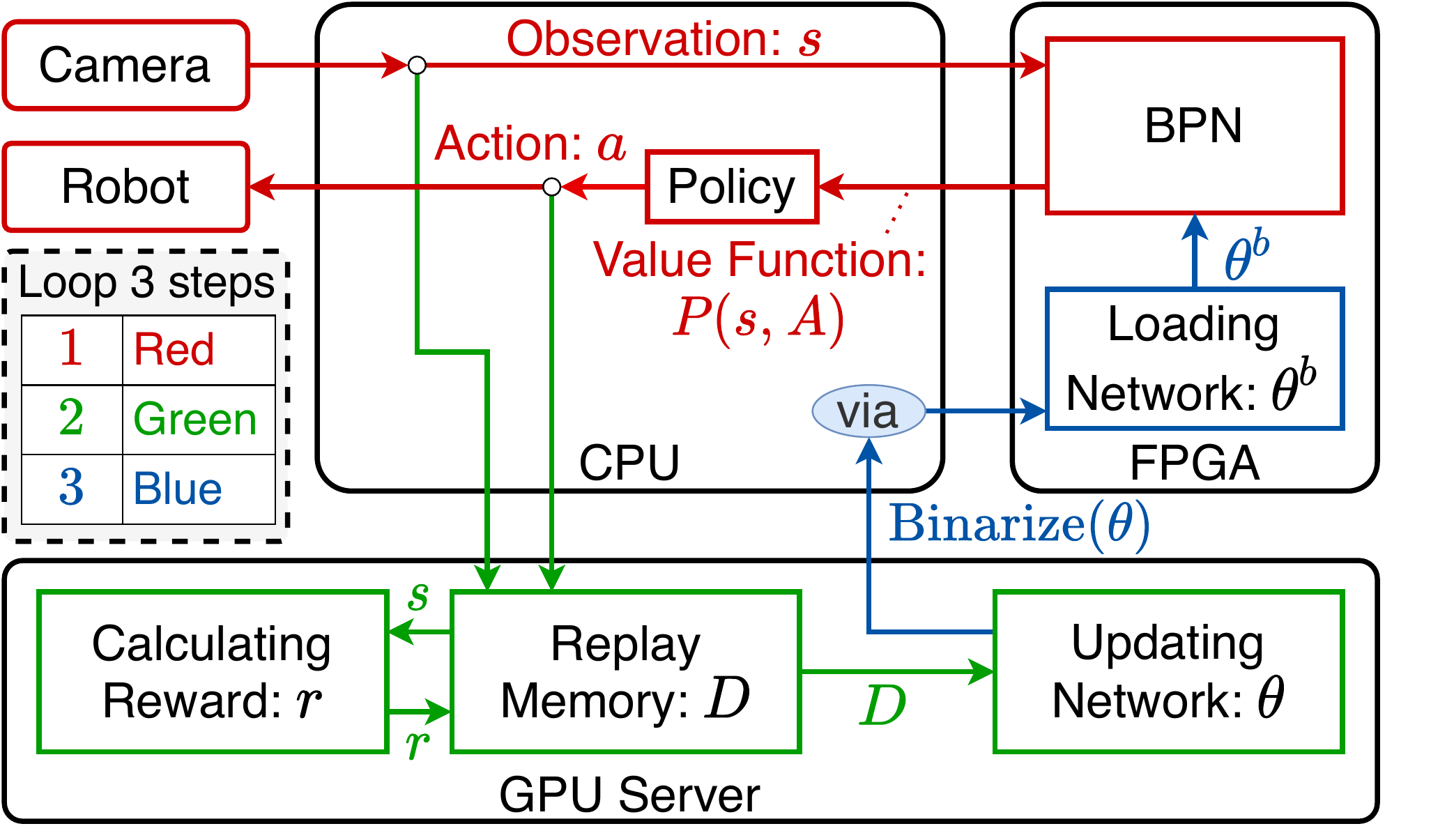}
                \caption{
                    Implemented learning system: Learning procedure consists of three steps.
                    (\textbf{1}) FPGA and CPU control robot to sample data $s$, $a$ into $\mathcal{D}$.
                    (\textbf{2}) $\mathcal{D}$ is used by GPU server to update BPN network parameters $\bs{\theta}$.
                    (\textbf{3}) FPGA updates BPN's network parameter $\bs{\theta}^b$.
                }
                \label{fig:learning-system}
            \end{center}
        \end{figure}

        \figref{fig:learning-system} shows a learning system that consists of three steps:
        (1) FPGA and CPU control the robot to collect datasets for learning.
        The CPU gets observation $s$ from a camera.
        The FPGA calculates action preference $P(s,\mathcal{A})$ from $s$.
        Then the CPU determines action $a$ based on the policy shown in \equref{eq:policy}.
        The robot executes $a$.
        (2) Based on the collected dataset in $\mathcal{D}$, network parameter $\bs{\theta}$ is updated based on \chapref{chap:BPN-learning}.
        Reward $r$ calculation is conducted on the GPU server instead of controlling the robot to maintain real-time control.
        (3) The GPU server binarizes and transfers the network parameter $\bs{\theta}$ to BRAM of the FPGA via the CPU.
        The FPGA calculates BPN using network parameter $\bs{\theta}^b$ loaded from BRAM.
        In this system, BRAM stores network parameters $\bs{\theta}^b$, which do not need to be compiled.
        The compiling time requires more than an hour.
        The DRL, which compiling network parameters $\bs{\theta}^b$ for updating $\bs{\theta}^b$ at each iteration $I$, has an extremely long learning time due to such a compilation time.

    \subsection{Learning Control Policies}
        \label{position_control}
        
        \subsubsection{Settings}
            \label{real-robot-ex-settings}
           The target task is the real-robot object tracking task shown in \figref{fig:senzai}.
            The tracking target, a blue marker, moves in a figure-8 pattern.
            The agent learns to keep the object in the camera frame.
            Separate robots, consisting of two servo motors (Dynamixel XM430-W350-T), control the agent and the target.
            The agent's motors are controlled by position-control and wait for converging them to the objective angle before taking the following control.
            The agent's initial position is fixed, and the target's initial position is randomly assigned within the range where the target is included in the camera frame.
            
            The observation is an RGB image of $84\times84$ pixels, as shown in \figref{fig:senzai}(Upper Left).
            As in \chapref{sim:tracking}, two consecutive frames are used as observation $s$.
            The motor rotation labels are $n=[-4,-2,0,2,4](\text {degree})$, and action $a$ is defined as all the combinations of $(0,n),(n,0),(-n,n),(n,-n)$ for two motor rotation angles $(\phi_{1}^{\text {agent}},\phi_{2}^{\text {agent}})$.
            The number of actions is $|\mathcal{A}|=17$.
            The trajectory of rotation angle $(\phi_{1}^{\text{target}},\phi_{2}^{\text{target}})$ of the two motors manipulating the target, with angular velocity $\omega$ and time step $t$, is $\phi_{1}^{\text{target}} = 25 \sin{\omega t}\ (\text{degree})$, and $\phi_{2}^{\text{target}} = 15 \sin{2 \omega t}\ (\text{degree})$.
            The definitions of reward and episode are identical as in \chapref{sim:tracking}.
            The network structure is same as in simulation tasks except for $N=100$.
            The learning parameters are different from those of \tabref{table:bpn_setting}: $\alpha=0.95$, $\beta=3$, $I=150$, $T=80$.
                
            \subsubsection{Results}
\begin{figure}[t]
                    \vspace{1.mm}
                    \centering
                    \begin{tabular}{c}
                        \hspace{-6mm}
                        \begin{minipage}{0.5\linewidth}
                            \centering
                            \subfigure[Learning Curves]{
                            \label{fig:real_result_learning_curve}
                            \includegraphics[keepaspectratio, height=35mm, angle=0]
                                        {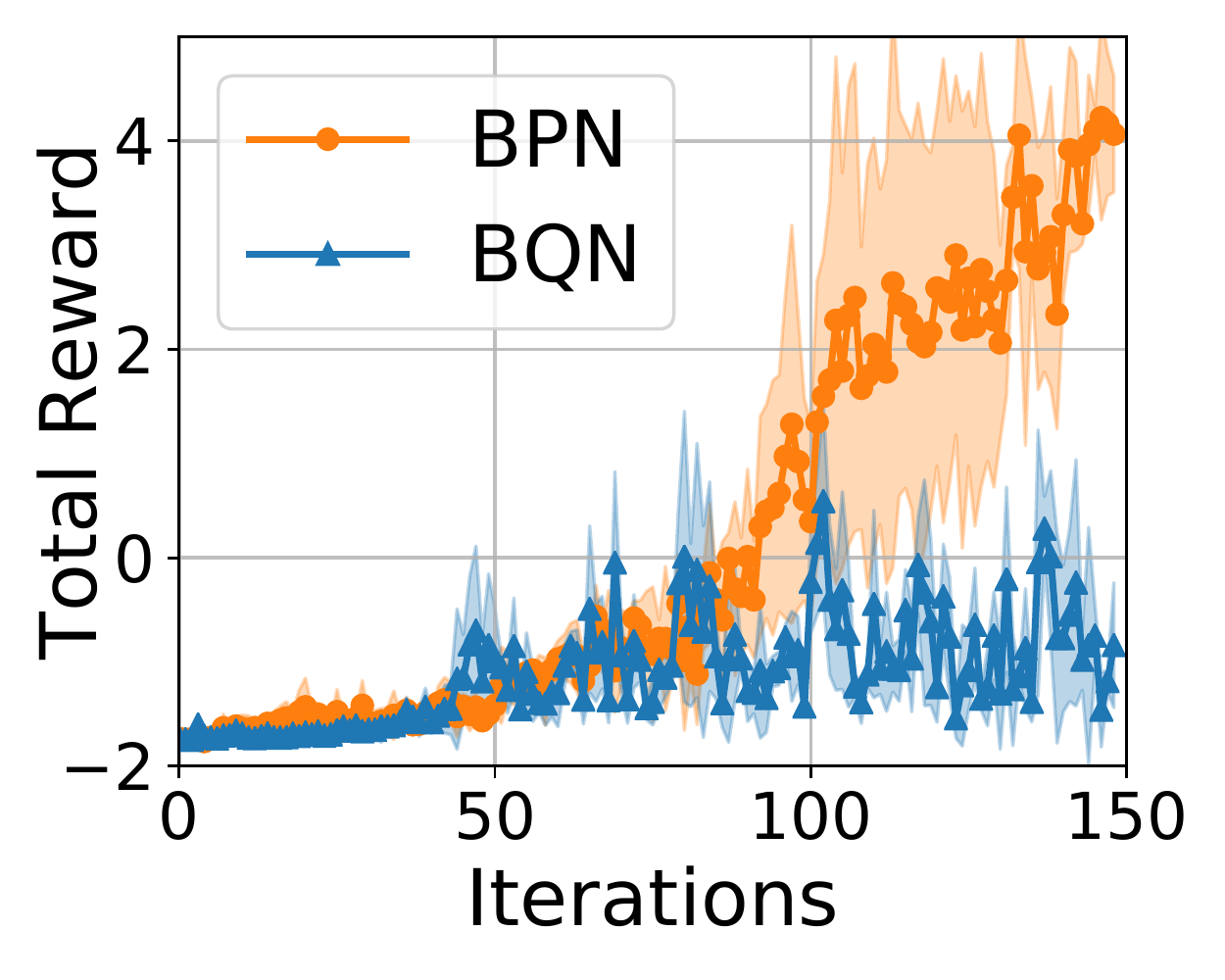}}
                        \end{minipage}
                        \hspace{1mm}
                        \begin{minipage}{0.5\linewidth}
                            \centering
                            \subfigure[Trajectories]{
                            \label{fig:real_result_trajectory_line}
                            \includegraphics[keepaspectratio, height=35mm, angle=0]
                                        {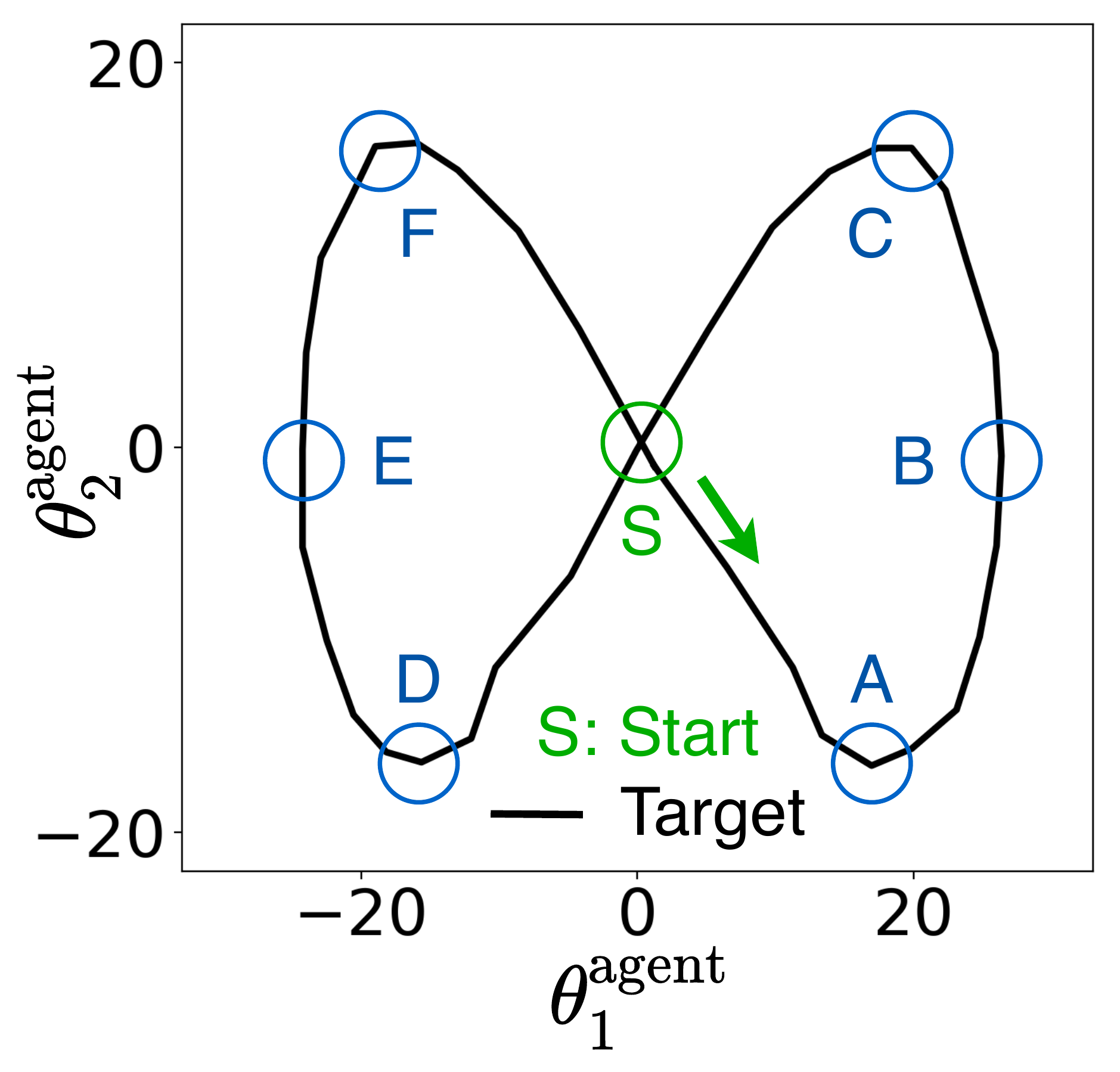}}
                        \end{minipage}
                        \\
                        \hspace{-7mm}
                        \begin{minipage}{1\linewidth}
                            \centering
                            \subfigure[Observations]{
                            \label{fig:real_result_trajectory_image}
                            \includegraphics[keepaspectratio, width=\linewidth, angle=0]
                                        {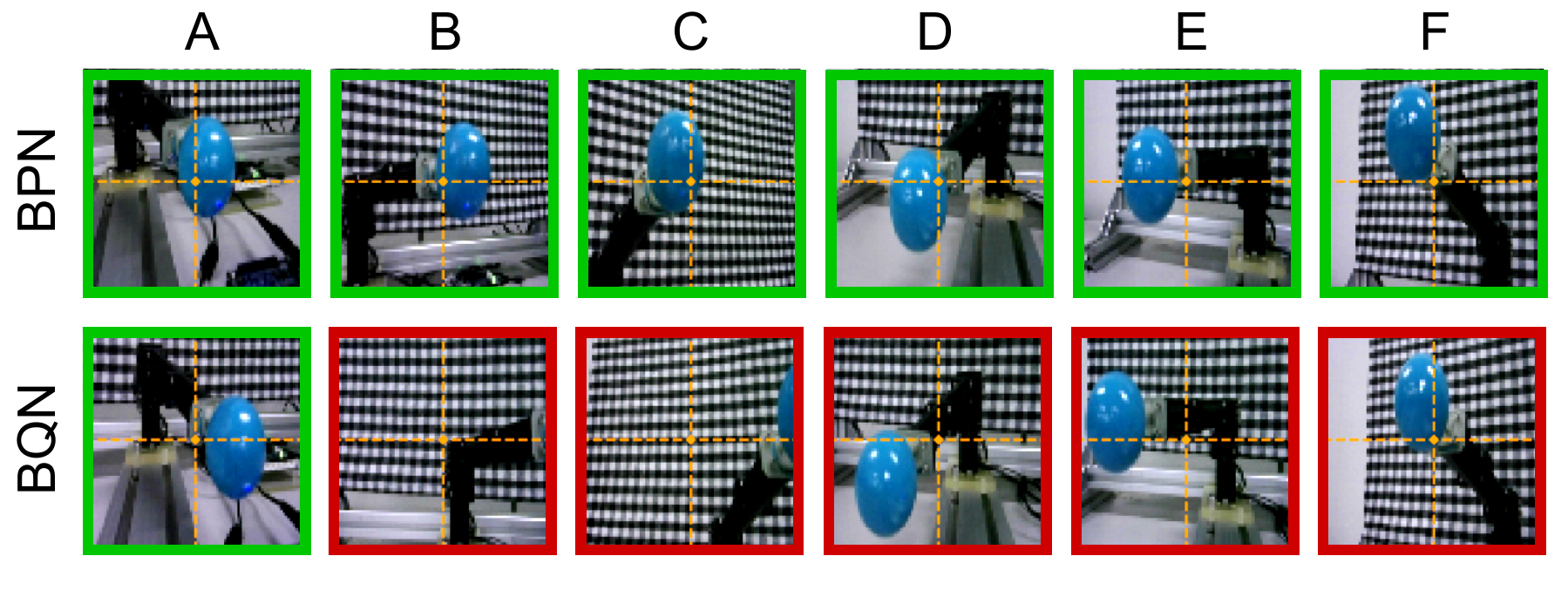}}
                        \end{minipage}
                    \end{tabular}
                    \caption{
                        Learning results of real-robot tracking task:
                        (\textbf{a})
                        Learning curves plot mean and variance of total reward per iteration $I$ over five experiments.
                        Entire run time of \algref{alg:BPN} is approximately 12 hours per experiment in $I=150$.
                        (\textbf{b}) 
                        Target trajectories of two motors of agent, $\theta_{1}^{\text{agent}}$ and $\theta_{2}^{\text{agent}}$, as they complete figure-8 pattern.
                        They are agent's motor trajectories when the target is at the center of the camera frame.
                        (\textbf{c})
                        Observations obtained from learned policy. 
                        A to F mean observed timing shown in (\textbf{b}).
                        Green and red frames indicate tracking success and failure.
                    }
                    \label{fig:real_result-all}
                \end{figure}

                The learning results are shown in \figref{fig:real_result-all}.
                \figref{fig:real_result_learning_curve} shows that the BQN did not learn progressively, although the BPN did.
                
                \figref{fig:real_result_trajectory_line} shows the target trajectory of the agent for each motor $\theta_{1}^{\rm agent}$ and $\theta_{2}^{\rm agent}$.
                \figref{fig:real_result_trajectory_image} is an observation of the learned policy when the target is in the agent's target trajectory A to F in \figref{fig:real_result_trajectory_line}.
                The BPN can track the target to fit in the camera frame using raw images as input in a natural background environment.
                BPN can also track the target in real-time without being delayed by the target.
                However, BQN is out of the frame from point B.
                
                The tracking time comparison between BPN and BQN is shown in \tabref{table:track_time}.
                BPN can track a target until the task end, which is six times longer than BQN.
                From \figref{fig:real_result_learning_curve}, the BQN does not learn a suitable policy.
                Thus, the BQN achieved tracking only for 1.9 seconds up to around the agent's target trajectory A in \figref{fig:real_result_trajectory_line}.
                
                \begin{table}
                    \vspace{1.mm}
                    \begin{center}
                    \caption{
                        Duration time of successful tracking of Real-Robot Experiment:
                        Control period of agent is 145 ms consisted of 141 ms for sampling two consecutive camera images and controlling robot's motor positions, and 4 ms inference time for NN in the FPGA.
                        The maximum step number is $T=80$ and corresponds to 11.6 s (145 ms $\times$ 80 steps).
                        Each time is the average of ten trials. 
                        \label{table:track_time}
                    }
                        \begin{tabular}{cccc}
                                    \toprule
                                        \textbf{Tracking-Time} & \textbf{RANDOM} & \textbf{BQN} & \textbf{BPN} \\
                                    \midrule
                                        \textbf{Second (Percent)} & 0.9 (8) & 1.9 (16) & 11.6 (100) \\
                                    \bottomrule
                        \end{tabular}
                    \end{center}
                \end{table}

    \subsection{Calculation Speed of FPGA}
        We verified that BPN can be implemented in edge FPGAs calculated in real-time by implementing BPN and DQN networks to FPGA.
        BPN's network (implemented in BCNN) is mainly calculated in logical operations.
        DQN's network (implemented in CNN) is mainly calculated in floating-point operations.
        In other words, we confirmed that BPN, implemented in logical operations for FPGA, has better hardware performance because its calculation speed is faster than DQN, which is implemented in floating-point operations.
        We implemented the networks on an FPGA evaluation board (Avnet Ultra96-V2) and verified the inference time.
        The network structures are same as \chapref{real-robot-ex-settings}.
        \tabref{table:resource}, which shows the results of implementing the networks, indicates that DQN cannot be applied to tasks that require fast calculation and that BPN can be applied to real-time control policies.

        \newcommand{\resourceTableWidth}{\hspace{4.0mm}}
        \begin{table}[t]
            \begin{center}
                \caption{
                    Network inference time:
                    Function approximation networks of BPN and DQN are implemented in FPGA evaluation board (Avnet Ultra96-V2) and
                    within it's resource capacity.
                    Network inference time is evaluated by Xilinx Vivado-HLS.
                }
                \label{table:resource}
                \begin{tabular}{ccc}
                    \toprule
                        \textbf{Calculation-Time per Inference} & \resourceTableWidth \textbf{DQN} & \resourceTableWidth \textbf{BPN} \\
                    \midrule
                        \textbf{ms / inference} & \resourceTableWidth 1003 & \hspace{2.0mm} 4 \\
                    \bottomrule
                \end{tabular}
            \end{center}
        \end{table}

        \label{communication_delay}
        We verify that the calculation of BPN on edge FPGAs is faster than that on other computers.
        The BPN has the same structure as \chapref{real-robot-ex-settings}.
        The experimental results of the calculation latency are shown in \tabref{table:communication_delay}.
        The latency is the highest when using a server.
        The latency is the lowest when using the edge FPGA and is reduced to less than 20\% of that of others. 
        
        \begin{table}
            \vspace{1.mm}
            \begin{center}
            \caption{
                Latency for getting actions from observations:
                We measure the time from receiving an image from a camera sensor to a control input to the edge robot.
                The calculation time is measured in the server as Intel Core i9-9900KS, and the edge CPU and FPGA as ARM Cortex-A53 and Xilinx ZU3EG A484 on Avnet Ultra96-V2, respectively. 
                All calculators get actions by calculating NNs (\textbf{Step 2}).
                The server has additional calculation steps, which are sending an image from the edge robot to the server (\textbf{Step 1}) and sending a control input from the server to the edge robot (\textbf{Step 3}).
                \label{table:communication_delay}
            }
                \begin{tabular}{lccc}
                    \toprule
                        \textbf{Steps} & \textbf{Server} & \textbf{Edge-CPU} & \textbf{Edge-FPGA} \\
                    \midrule
                        \textbf{1. Send Images} & 34 ms & - & - \\
                        \textbf{2. Calculate NNs} & 2 ms & 21 ms & 4 ms \\
                        \textbf{3. Send Controls} & 2 ms & - & -    \\
                    \midrule
                        \textbf{Total-Time} & 38 ms & 21 ms & 4 ms \\
                    \bottomrule
                \end{tabular}
            \end{center}
        \end{table}

\section{Discussions}
    \label{Discussions}
    
    \chapref{ex:real} shows how to learn an object tracking task by DRL with a real robot and FPGA.
    An extension of this work might apply autonomous edge-robot control to exploit FPGAs' power-saving nature. 
    To build a DRL system for such a purpose, we need an environment where FPGA agents can communicate with a server that updates the control policies. 
    
    \chapref{chap:real_ex_sistem} suggests that learning by an autonomous robot requires a communication environment between the FPGA and the server. 
    A bottleneck in implementing a learning algorithm on FPGAs is implementing a large-scale, error back-propagation (BP) algorithm, which might be addressed with \cite{DFA}. 
    In addition, extending the BP implementation method for servers \cite{FPGA-drl-FPGAandGPUforBP-a3c,fpga-drl-BPinServer} may give some tips for implementing the BP fast in edge FPGAs.

    The BPN shown in Section IV is a learning method that assumes a discrete action space. However, continuous actions are often required in robot control tasks. 
    The extension of the proposed method to continuous action space remains our future work. To this end, we could adopt the actor-critic architecture \cite{sac}; however, we need to be concerned about how to represent the actor and critic accurately with BCNNs, which have low accuracy in function approximation. 
    
\section{Conclusion}
    \label{Conclusion}
    We proposed a Binarized P-Network as a DRL algorithm suitable for FPGAs. We also implemented the BPN for an object tracking task with a real robot using image inputs and confirmed its effectiveness.

\addtolength{\textheight}{-0cm}   


\bibliographystyle{IEEEtran}
\bibliography{IEEEabrv,mybibfile}

\end{document}